\documentclass[10pt,conference,letterpaper]{ieeeconf}
\IEEEoverridecommandlockouts

\usepackage{amsmath,amssymb,amsfonts}
\usepackage{bm}
\usepackage{graphicx}
\usepackage{algorithm,algorithmic}
\usepackage[dvipsnames,svgnames]{xcolor}
\usepackage{paralist}

\newcommand{\x}{\bm{x}}
\newcommand{\q}{\bm{q}}
\newcommand{\z}{\bm{z}}
\newcommand{\e}{\bm{e}}
\newcommand{\cov}[3]{\bm{k}_{#1,#2}^{(#3)}}
\newcommand{\ucb}{UCB}
\newcommand{\qed}{$\blacksquare$}
\newcommand{\real}{\mathbb{R}}
\newcommand{\ar}{\mbox{AR-1}}

\newtheorem{theorem}{Theorem}[section]
\newtheorem{proposition}{Proposition}[section]
\newtheorem{remark}{Remark}[section]

\newtheorem{lemma}{Lemma}[section]

\graphicspath{{./images/}}

\begin{document}
\title{On Multi-Fidelity Impedance Tuning for Human-Robot Cooperative Manipulation \thanks{This work was supported in part by ARO grant W911NF-18-1-0325 and in part by NSF Award CNS-2134076.}}
\author{Ethan Lau \and Vaibhav Srivastava \and Shaunak D.~Bopardikar \thanks{The authors are with the Electrical and Computer Engineering Department at Michigan State University.}}

\maketitle

\begin{abstract}
We examine how a human-robot interaction (HRI) system may be designed when input-output data from previous experiments are available. In particular, we consider how to select an optimal impedance in the assistance design for a cooperative manipulation task with a new operator. Due to the variability between individuals, the design parameters that best suit one operator of the robot may not be the best parameters for another one. However, by incorporating historical data using a linear auto-regressive (\ar) Gaussian process, the search for a new operator's optimal parameters can be accelerated. We lay out a framework for optimizing the human-robot cooperative manipulation that only requires input-output data. We establish how the \ar\ model improves the bound on the regret and numerically simulate a human-robot cooperative manipulation task to show the regret improvement. Further, we show how our approach's input-output nature provides robustness against modeling error through an additional numerical study.
\end{abstract}

\section{Introduction}
Recently, there has been an expansion of robotic automation across many industries. Industrial robots exceed humans in strength and precision, and they can successfully perform structured, repetitive tasks. However, increasingly complex tasks require increasingly complex robots. Situations often arise in which a robot cannot complete a task on its own.
By bringing a human into the loop, HRI leverages a human's perceptive and decision-making strengths while still benefiting from the robot's precision or physical strength.

A common robot found in HRI is the robotic manipulator, a multi-segmented arm that accomplishes tasks using its end-effector. Using an impedance model, a manipulator's interaction with the environment is often controlled by adjusting its effective mass, stiffness, and damping at its end-effector
\cite{hogan1985impedance}. The impedance model simplifies control strategies by dynamically relating the manipulator's position and force. 
Multiple types of impedance control methods have been proposed, including adaptive control \cite{lu1991impedance, huo2021model,sun2023repetitive}, iterative methods \cite{li2018iterative}, and neural networks \cite{yang2018neural}. Studies have also analyzed variable impedance models \cite{ficuciello2015variable} and their stability \cite{sun2019stability}.

Robotic manipulators have found many engineering applications, including exosuits \cite{li2022human} and construction automation \cite{bock2016construction}. We specifically consider a cooperative manipulation task in which a human works with a manipulator to track a large object along a given trajectory. The manipulator seeks to follow a general trajectory but requires the human to provide an auxiliary force to guide the object's path. In this context, the human can be modeled using a transfer function specified by a set of gains \cite{yang2021data, modares2015optimized, li2017adaptive}. These gains may vary between individuals, resulting in a specialized tuning for each operator.
As a result, a trade-off is encountered when a new operator must be trained.
In a purely robotic setting, the system structure may be found using system identification; however, this process may prove time consuming and annoying for the operator, leading to operator impatience. Iteratively tuning the system for the new operator would also waste time and valuable historical data. Meanwhile, solely relying on historical data may result in suboptimal performance.
Our goal is to leverage previous operator data while finding the ideal tuning parameters for a new operator.

To do so, we use Gaussian process (GP) regression, a tool commonly used to model and optimize unknown and difficult-to-evaluate cost functions \cite{williams2006gaussian}. One benefit of GPs is their inclusion of confidence bounds in their prediction. Multi-fidelity Gaussian processes (MF-GP) use multiple correlated inputs to predict an output. Specifically, the \ar\ model relates data across various inputs through a nested linear structure. 
\ar\ models have been used to incorporate low-fidelity data from a simulation in order to optimize a high-fidelity function related to the true system \cite{marco2017virtual, lau2023multi}.


The following are our main contributions:
\begin{compactenum}
\item Using an impedance controller for the robotic manipulator and a transfer function model for human input, we formulate the optimal assistance design for cooperative manipulation as an input-output problem where the system gains are the inputs and the system performance is the output. By applying a Gaussian process framework to this problem, we develop a sequential method to find the system's optimal gains that requires \emph{only this input-output data}.
\item We incorporate previous operators' input-output data through the use of a multi-fidelity Gaussian process. By analytically quantifying how multi-fidelity affects the conditional covariance, we provide an upper bound on the regret. Additionally, we relate this bound to the measurement quality and variability across operators to show that an increase in the accuracy of prior data leads to decrease in the regret. 
\item We numerically simulate input-output data for a model of human-robot cooperative manipulation in order to compare the single- and multi-fidelity formulations. We provide an example where cumulative and best instantaneous regret is better for the multi-fidelity formulation than the single-fidelity formulation. Further, we simulate a disturbance-impacted model of the human-robot manipulator to demonstrate the robustness of our approach. 

\end{compactenum}

\begin{figure*}
    \centering    \includegraphics[width=0.7\linewidth]{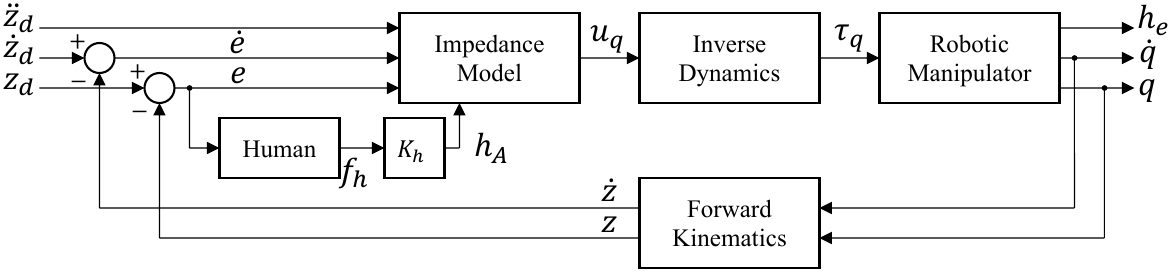}
    \caption{Block Diagram of the Human-Robot Manipulator System.}
    \label{fig:blockdiagram}
\end{figure*}

\section{System Description}
Consider a cooperative manipulation system, in which a human and robot seek to maneuver on object along a given trajectory. 
The human may be required to exert some effort (e.g. by lifting the object) but the robot can seek to assist the human in other ways (e.g. through precise maneuvering). Given the object's position, both the human and robot know the tracking error and can take a control action based on the error and desired trajectory information.

In this section, we formulate a model for this cooperative manipulation system. 
In general, robotic manipulators are nonlinear, but using feedback linearization, we design a control input so that the robot behaves as an impedance model. The impedance model allows the human-robot system to be formulated as a linear time-invariant system, which can then be controlled using state feedback. An overview of this control strategy is displayed in Fig. \ref{fig:blockdiagram}.

\subsection{Robot Impedance Model}
Consider an $n$-link robot manipulator with the joint space dynamical model \cite{siciliano2010robotics}
\begin{align}\label{eq:eom_q}
    M_q(\q)\ddot{\q} {+} C_q(\q,\dot{\q})\dot{\q}{+}F_q\dot{\q}{+}G_q(\q) = \bm{\tau}_q {-} J^T(\q) \bm{h}_e,
\end{align}
where $\q \in \real^n$ is the manipulator's position in the joint space with $n$ degrees of freedom. Here, $M_q(\q) \in \real^{n\times n}$ is the symmetric positive definite inertia matrix, $C_q(\q,\dot{\q}) \in \real^{n\times n}$ is the Coriolis-centrifugal matrix, $F_q \in \real^{n\times n}$ is the vector of damping coefficients, $G_q(\q) \in \real^{n}$ is the vector of gravitational forces, $\bm{\tau}_q \in \real^{n}$ are the input torques at the joints, 
$\bm{h}_e \in \real^{n}$ are the contact forces exerted by the manipulator's end-effector, and $J \in \real^{n\times n}$ is the geometric Jacobian relating the end-effector velocities to the joint velocities.

Let $\z$ and $\z_d$ be the position and desired position of the manipulator end-effector. 
The error between these positions is given by 
\begin{align}
    \e := \z - \z_d.
\end{align}
Assuming the joint positions $\q$ and velocities $\dot{\q}$ are known, feedback linearization may be used to control the system.
We define a control law
\begin{align} \label{eq:tau_q}
    \bm{\tau}_q = M_q(\q)\bm{u}_q+p(\q,\dot{\q}) + J^T(\q)\bm{h}_e,
\end{align}
where
\begin{align}
    p(\q,\dot{\q}) = C_q(\q,\dot{\q})\dot{\q} + F_q\dot{\q}+G_q(\q).
\end{align}
Selecting $M_m$, $B_m$, and $K_m$ as the desired inertia, damping, and stiffness matrices of the impedance model, we set the input $\bm{u}_q$ of \eqref{eq:tau_q} to
\begin{align} \label{eq:u_q}
    \bm{u}_q =& J_{A}^{-1}(\q)M_m^{-1}\\
    &\times (M_m \ddot{\z_d} + B_m \dot{\e} + K_m \e - M_m \dot{J}_{A}(\q,\dot{\q})\dot{\q} - \bm{h}_A), \nonumber
\end{align}
where $J_{A}(\q)$ is the analytical Jacobian satisfying \mbox{$\dot{\z}_d = J_{A}(\q) \dot{\q}$}, and $\bm{h}_A$ is the forcing vector of the impedance model. 

Assume the forcing vector $\bm{h}_A$ takes the form
\begin{align} \label{eq:h_a}
    \bm{h}_A = K_h \bm{f}_h,
\end{align}
where $\bm{f}_h$ is the human control effort and $K_h\in \real^{n\times n}$ is a diagonal matrix of gains.
Then, combining \eqref{eq:eom_q}, \eqref{eq:tau_q}, \eqref{eq:u_q}, and \eqref{eq:h_a}, we obtain the impedance model
\begin{align} \label{eq:eom_e}
    M_m \ddot{\e} + B_m \dot{\e} + K_m \e = K_h \bm{f}_h.
\end{align}

Define the augmented error vector as $\overline{\e}:=[\e^T\ \dot{\e}^T]^T \in \real^{2n}$. Then \eqref{eq:eom_e} can be rewritten as
\begin{align}
    \dot{\overline{\e}} = A \overline{\e} + B \overline{\bm{u}},
\end{align}
where
\begin{align}
	A =
	\begin{bmatrix}
		\bm{0} & I_n \\
		\bm{0} & \bm{0}
	\end{bmatrix}\in \real^{2n \times 2n}, \quad
     B =
    \begin{bmatrix}
        \bm{0}\\
        I_{n}
    \end{bmatrix} \in \real^{2n \times n},
\end{align}
and
\begin{align} \label{eq:u_bar}
    \overline{\bm{u}} = -M_m^{-1} \left[ K_m\ B_m \right] \overline{\e} + M_m ^{-1}K_h \bm{f}_h.
\end{align}

\subsection{Human Impedance Model}
To account for the effect of the human in the HRI system, we model the human operator using a proportional gain and a derivative gain \cite{yang2021data}. Assuming the human's reaction is based on the robot error $\e$, we obtain the human impedance model
\begin{align} \label{eq:human_ss}
    K_d \dot{\bm{f}}_h + K_p \bm{f}_h  = \bm{e},
\end{align}
where $K_d, K_p \in \real^{n\times n}$ are diagonal matrices of human gains. These gains are considered to be unknown and may vary between operators. As such, we denote by $K_d^i$ and $K_p^i$ the gain matrices of the $i$-th operator.
Using these operator-specific gains, \eqref{eq:human_ss} can be rewritten as
\begin{align}
    \dot{\bm{f}}_h = A_h^i \bm{f}_h + B_h^i \e,
\end{align}
where 
\begin{align}
    A_h^{i} &= -[K_d^{i}]^{-1} K_p^i &&\in \real^{n\times n},\\
    B_h^{i} &= \left[ [K_d^{i}]^{-1} \quad  \bm{0}\right]  &&\in \real^{n\times 2n}.
\end{align}

\subsection{Human-Robot Impedance Model}
With models established for the robot and human, we now write an augmented state space model for the system. Define the augmented state as $Z := [\bar{\e}^T, \bm{f}_h^T]^T \in \real^{3n}$. Then the HRI manipulator for the $i$-th operator has the state space model
\begin{align} \label{eq:Z_i_dot}
    \dot{Z}_i = \mathcal{A}^{i}Z_i + \mathcal{B}^{i} \bm{u},
\end{align}
where 
\begin{align} \label{eq:augmented_state_space}
	\mathcal{A}^{i} =
	\begin{bmatrix}
		A & \bm{0} \\
		B_h^{i} & A_h^{i}
	\end{bmatrix} \in \real^{3n\times3n},
    \quad
    \mathcal{B}^{i} =
	\begin{bmatrix}
		B \\
		\bm{0}
	\end{bmatrix} \in \real^{3n\times n},
\end{align}
and
\begin{align} 
\label{eq:u_e}
    \bm{u} = -K Z_i,
\end{align}
with control gains
\begin{align} \label{eq:K}
	K =
    M_m^{-1}
	\begin{bmatrix}
		K_m & B_m & K_h
	\end{bmatrix} \in \real^{n\times 3n}.
\end{align}

Given a set of control gains $K$, the quadratic cost of cooperative manipulation for the $i$-th operator is
\begin{align} \label{eq:quad_cost}
    J_i(K) &= \int_{0}^{\infty} (Z_i^T(\tau) Q Z_i(t) + \bm{u}^T(\tau) R \bm{u}(\tau)) d\tau \\
    &=\int_{0}^{\infty} Z_i^T(\tau) [Q + K^T R K] Z_i(t) d\tau,
\end{align}
where $Q \in \real^{3n\times 3n}$ weights the effect of the tracking error, error rate, and human effort, $R \in \real^{n\times n}$ weights the effect of the robot's control effort, and 
$Z_i(\tau)$ is the solution of \eqref{eq:Z_i_dot} given an initial condition $Z_i(0)$ and feedback controller \eqref{eq:u_e}.

\subsection{Problem Statement}
Consider an HRI system with the impedance model \eqref{eq:augmented_state_space}. Let $K(\x)$ be a controller depending on design parameters $\x \in \mathcal{X} \subset\real^q$. 
Suppose that the robot has $m$ human operators, with the $i$-th human possessing their own performance metric 
\begin{align} \label{eq:f_i}
    f_i(\x) = -J_i\left(K(\x)\right).
\end{align}
As the $i$-th operator tests different design parameters, they obtain data for $\bm{X}_i \subseteq \mathcal{X}$.

Now, suppose a new $(m{+}1)$-th human operates the same robot. 
Our goal is to leverage the previous data $\left(\bm{X}_i, f_i(\bm{X}_i)\right)$ to find an ideal set of design parameters $\x^*$ that optimizes the new operator's performance $f_{m+1}$.

\section{Using Previous Data in Multi-Fidelity Methods for Control Gain Selection}
With our problem statement established, we provide an overview of Gaussian processes. We introduce the notion of multi-fidelity and describe how the HRI problem is formulated to fit this framework.
\subsection{Gaussian Processes (GPs)}
A Gaussian process is a collection of random variables, in which any finite subset of variables has a multivariate Gaussian distribution \cite{williams2006gaussian}. A GP is defined by its mean function $\mu(\x)$ and its covariance (kernel) function $k(\x, \x')$. 

For a set of inputs $\bm{X}_t = \{\x_1,\dots, \x_t\}$, we can create a covariance matrix $\bm{k}(\bm{X}_t, \bm{X}_{t}) = [k(\x_i,\x_j)]_{i,j=1}^{t,t}$. By taking the covariance between a point and a set of points, we obtain a covariance vector $\bm{k}(\x) := \bm{k}(\bm{X}_t,\x) = [k(\x_1, \x) \ldots k(\x_t,\x)]^T$.

Let $\bm{Y}_t = [y_1,\dots, y_t]^T$ be noisy samples of $f$ at $\bm{X}_t$, where $y_i = f(\x_i) +\eta$ has independent and identically distributed Gaussian measurement noise $\eta\sim N(0,\xi^2)$.

Then the posterior distribution of $f$ is another GP with mean $\mu_{t+1}$, covariance $k_{t+1}$, and standard deviation $\sigma_{t+1}$ given by
\begin{align}
    \mu_{t+1}(\x) &=
    \bm{k}^T(\x)
    [\bm{k}(\bm{X}_t,\bm{X}_t) + \xi^2 I]^{-1} \bm{Y}_t, \label{eq:gp_mean} \\
    k_{t+1}(\x,\x') &= k_t(\x,\x'){-}\bm{k}^T(\x)
    [\bm{k}(\bm{X}_t,\bm{X}_t){+}\xi^2 I]^{-1} \bm{k}(\x'), \nonumber \\
    \sigma_{t+1}(\x) &= \sqrt{k_t(\x,\x)}. \label{eq:gp_std}
\end{align}

In problems where a GP is being optimized, Bayesian optimization is an iterative framework used to select the next point to evaluate. Popular Bayesian optimization approaches include using the Expected Improvement \cite{jones1998efficient} and the Upper Confidence Bound (\ucb) \cite{srinivas2012information}.

The \ucb\ algorithm selects points according to 
\[
\x_t = \underset{\x \in \mathcal{X}}{\arg\max}\ \mu_{t-1}(\x) + \beta_t^{1/2}\sigma_{t-1}(\x),
\]
where $\beta_t$ is a parameter which controls the algorithm's tendency to explore. This algorithm is formalized in Alg. \ref{alg:ucb}. One particular appeal of \ucb\ are its theoretical guarantees associated with a metric called regret.

\begin{algorithm} [tbh] 
\caption{UCB Sampling} \label{alg:ucb}
\begin{algorithmic}[1]
    \STATE \textbf{Input:} GP $f$ with priors $\mu_{0}$, $\sigma_{0}$, Discrete domain $\mathcal{X}$
    \FOR{$t=1,2,\dots$}
    \STATE Choose $\x_t = \underset{\x\in \mathcal{X}}{\arg\max\ }\mu_{t-1}(\x)+\beta_t^{1/2}\sigma_{t-1}(\x)$
    \STATE Sample $y_t(\x_t) = f(\x_t) + \eta$
    \STATE Predict $\mu_{t}(\x)$, $\sigma_{t}(\x)\ \forall \x\in \mathcal{X}$
    \ENDFOR
\end{algorithmic}
\end{algorithm}

For an iterative optimization algorithm, the instantaneous regret of an evaluation is given by 
\begin{align} \label{eq:instantaneous_regret}
    r_t(\x_t) =  f(\x^*) - f(\x_t),
\end{align}
where $\x^* = \underset{\x \in \mathcal{X}}{\arg\max}\ f(\x)$. Regret indicates the gap between the current evaluation and the best possible evaluation. After $T$ rounds, the cumulative regret is given by $R_T = \sum_{t=1}^T r_t$ and the best instantaneous regret is given by $r^*_T=\min_{t=\{1...T\}} r_t$.

\subsection{Multi-Fidelity Gaussian Processes (MF-GPs)}
An MF-GP incorporates data from multiple inputs to model $f$. One type of MF-GP is the \ar\ model \cite{kennedy2000predicting}. \ar\ models $f$ as a linear combination of a low-fidelity GP $f_L(\x)$ and an error GP $\delta(\x)$ by
\begin{align} \label{eq:ar1}
	f(\x) = \rho f_L(\x) + \delta(\x),
\end{align}
where $\rho$ is a scaling constant.

Denote the kernels of $f_L$ and $\delta$ by $\bm{k}^{(L)}$ and $\bm{k}^{(\delta)}$, respectively, and let evaluations of $f_L$ and $f$ have variances $\xi_L^2$ and $\xi_H^2$.
Then, for $\bm{X} = [\bm{X}_L, \bm{X}_H]$, an \ar\ model has a covariance matrix of the form
\begin{align} \label{eq:covariance_matrix}
    \bm{k}^{(MF)}(\bm{X},\bm{X})=
    \begin{bmatrix}
        \cov{L}{L}{L} {+} \xi_L^2 I & \rho \cov{L}{H}{L} \\
        \rho \cov{H}{L}{L} & \rho^2 \cov{H}{H}{L} {+} \cov{H}{H}{\delta} {+} \xi_H^2 I \\
    \end{bmatrix},
\end{align}
where $\cov{H}{L}{L}$ is shorthand notation for the single-fidelity covariance matrix $\bm{k}^{(L)}(\bm{X}_H,\bm{X}_L)$. 

Unlike larger GP models, the \ar model allows for the iterative updating of each fidelity, thereby maintaining a computational complexity on the same order as a single-fidelity GP. Additionally, it's decoupled recursive structure allows for the computationally efficient learning of its parameters.

\subsection{Multi-Fidelity Approach to Control Design}
Using the \ar\ model, we aim to effectively leverage data from the previous operators to a specific individual. Consider a set of $m+1$ operators, with the $i$-th operator's performance data $\left(\bm{X}_i, f_i(\bm{X}_i)\right)$.

Let $f: \mathcal{X}\to \real$ be an unknown realization of a GP with \ar\ structure \eqref{eq:ar1}. 
Because the quadratic cost is sufficiently smooth with respect to $K$, we assume the GP $f$ adequately represents the performance $f_{m+1}$ of the $(m{+}1)$-th operator.
Meanwhile, we treat $f_L$ as a GP with observations the first $m$ operators. Note, $f_L$ does not specifically represent any $f_i$ but rather models the expected performance of the previous $m$ operators. 

Using \ucb, we iteratively select an $\x_t$ to test for the $(m{+}1)$-th operator, thereby obtaining evaluations of $f$.
This Multi-Fidelity Formulation (MFF) is formalized in Algorithm \ref{alg:mf}.

\begin{algorithm} [tbh] 
\caption{Multi-Fidelity (MFF) Formulation} \label{alg:mf}
\begin{algorithmic}[1]
    \STATE \textbf{Input:} Data $\left(\bm{X}_i, f_i(\bm{X}_i)\right)$ for $i\in \{1, 2,\dots,m+1\}$, Discrete domain $\mathcal{X}$
    \STATE Let $f_L$ be a GP with evaluations $\left(\bm{X}_i, f_i(\bm{X}_i)\right)$ for $i=\{1,\dots,m\}$  
    \STATE Let $f$ be a GP with form $f(\x) = \rho f_L(\x) + \delta(\x)$ and evaluations $\left(\bm{X}_{m+1}, f_{m+1}(\bm{X}_{m+1})\right)$
    \STATE Predict $\mu_{0}(\x)$, $\sigma_{0}(\x)\ \forall \x\in \mathcal{X}$
    \STATE UCB($f, \mu_{0}, \sigma_0, \mathcal{X}$) 
\end{algorithmic}
\end{algorithm}

We compare MFF to two single-fidelity approaches that do not take advantage of the \ar\ structure. In the Collective Single-Fidelity (CSF) Formulation of Algorithm \ref{alg:csf}, data from all operators is treated as a single fidelity. In the Limited Single-Fidelity (LSF) Formulation of Algorithm \ref{alg:lsf}, the single-fidelity GP contains only data from the new $(m{+}1)$-th operator. Essentially, LSF is a naive approach that ignores any previous operator data.
\begin{algorithm} [th] 
\caption{Collective Single-Fidelity (CSF) Formulation} \label{alg:csf}
\begin{algorithmic}[1]
    \STATE \textbf{Input:} Data $\left(\bm{X}_i, f_i(\bm{X}_i)\right)$ for $i\in \{1, 2,\dots,m+1\}$, Discrete domain $\mathcal{X}$
    \STATE Let $f$ be a GP with evaluations $\left(\bm{X}_i, f_i(\bm{X}_i)\right)$ for $i=\{1,\dots,m+1\}$  
    \STATE Predict $\mu_{0}(\x)$, $\sigma_{0}(\x)\ \forall \x\in \mathcal{X}$
    \STATE UCB($f, \mu_{0}, \sigma_0, \mathcal{X}$) 
\end{algorithmic}
\end{algorithm}

\begin{algorithm} [th] 
\caption{Limited Single-Fidelity (LSF) Formulation} \label{alg:lsf}
\begin{algorithmic}[1]
    \STATE \textbf{Input:} Data $\left(\bm{X}_{m+1}, f_{m+1}(\bm{X}_{m+1})\right)$, Discrete domain $\mathcal{X}$
    \STATE Let $f$ be a GP with evaluations $\left(\bm{X}_{m+1}, f_{m+1}(\bm{X}_{m+1})\right)$ 
    \STATE Predict $\mu_{0}(\x)$, $\sigma_{0}(\x)\ \forall \x\in \mathcal{X}$
    \STATE UCB($f, \mu_{0}, \sigma_0, \mathcal{X}$) 
\end{algorithmic}
\end{algorithm}

\section{Theoretical Results}
With the multi-fidelity nature of this problem established, we now examine how the properties of \ar\ GPs improve the regret performance of \ucb.
We start with a proposition used to calculate a bound on the conditional covariance.
\begin{proposition} \label{prop:matrix_ineq}
Let $Q$ be a positive definite matrix and $\sigma\in\real$ be any scalar such that $\sigma < \sqrt{\lambda_{min}(Q)}$. Then
\begin{align*}
    (Q+\sigma^2 I)^{-1} \succeq Q^{-1} - \sigma^2 Q^{-2}.
\end{align*}
\end{proposition}
\smallskip

Note, this is a specified form of \cite[Eq. (191)]{petersen2008matrix}, which denotes it as an approximation but does not state a direction of inequality. 

\smallskip
\begin{proof}
We rewrite
    \begin{align}
        (Q+\sigma^2 I)^{-1} &= (QQ^{-1}Q + \sigma^2 Q^{-1}Q)^{-1} \nonumber \\
        &= ((I+\sigma^2 Q^{-1})Q)^{-1}  \nonumber\\
        &= Q^{-1} (I+\sigma^2 Q^{-1})^{-1}. \label{propeq1}
    \end{align}
    By writing the series expansion of the second factor,
    \begin{align*}
        (I{+}\sigma^2 Q^{-1})^{-1} &= I {-} \sigma^2 Q^{-1} {+} (\sigma^2 Q^{-1})^2 {-} (\sigma^2 Q^{-1})^3 {+} \dots \nonumber\\ 
        &=  I {-} \sigma^2 Q^{-1} {+}\sigma^2 Q^{-1}(I {-} \sigma^2 Q^{-1}+...) Q^{-1} \nonumber \\
        &=  I {-} \sigma^2 Q^{-1} {+}\sigma^2 Q^{-1}(I{+}\sigma^2 Q^{-1})^{-1} Q^{-1}. \nonumber
    \end{align*}
    Since $(I{+}\sigma^2 Q^{-1})^{-1}$ and $Q^{-1}$ are positive definite, their product is positive definite and 
    \begin{align}
        (I{+}\sigma^2 Q^{-1})^{-1}&\succeq I - \sigma^2 Q^{-1}. \label{propeq2}
    \end{align}
    By substituting \eqref{propeq2} into \eqref{propeq1}, we complete the proof.
\end{proof}
\smallskip

\begin{lemma}[Cond. Covariance of a Noisy \ar\ GP]\label{thm:conditional_covariance}
    Consider an \ar\ GP with high-fidelity evaluations at $\bm{X}_H$ and low-fidelity evaluations at $\bm{X}_L$. For a sufficiently small $\xi_L^2$, the covariance of the high-fidelity data conditioned on the low-fidelity data can be upper bounded by $\tilde{\bm{k}}^{(MF)}$, where
    \begin{align*}
        \tilde{\bm{k}} :=& \rho^2 \cov{H}{H}{L} + \cov{H}{H}{\delta} + \xi_H^2I - \rho^2 \cov{H}{L}{L} [\cov{L}{L}{L}]^{-1}\cov{L}{H}{L} \\
        & +\xi_L^2 \cov{H}{L}{L}[\cov{L}{L}{L}]^{-2} \cov{L}{H}{L}.
    \end{align*}
\end{lemma}
\smallskip
\begin{proof}
The conditional covariance of an \ar\ GP can be written as
    \begin{align*}
        &\bm{k}(f_H(\bm{X}_H),f_H(\bm{X}_H) |f_L(\bm{X}_L)=\bm{y}_L,f_H(\bm{X}_H)=\bm{y}_H) \nonumber\\
        &=
        \rho^2 \cov{H}{H}{L} + \cov{H}{H}{\delta} + \xi_H^2I - \rho^2 \cov{H}{L}{L} [\cov{L}{L}{L} + \xi_L^2I]^{-1}\cov{L}{H}{L}\nonumber\\
        &\preceq
        \rho^2 \cov{H}{H}{L} + \cov{H}{H}{\delta} + \xi_H^2I \nonumber  \\
        &\phantom{\leq}
        - \rho^2 \cov{H}{L}{L} \left( [\cov{L}{L}{L}]^{-1}{-}\xi_L^2 [\cov{L}{L}{L}]^{-2} \right) \cov{L}{H}{L} \nonumber \\
        &=
        \rho^2 \cov{H}{H}{L} + \cov{H}{H}{\delta} + \xi_H^2I - \rho^2 \cov{H}{L}{L} [\cov{L}{L}{L}]^{-1}\cov{L}{H}{L} \nonumber\\
        &\phantom{=} +\xi_L^2 \cov{H}{L}{L}[\cov{L}{L}{L}]^{-2} \cov{L}{H}{L},
    \end{align*}
    where the inequality is obtained from Proposition \ref{prop:matrix_ineq}.
\end{proof}
\smallskip

\begin{remark}
Recall, $f_L$ represents the expected performance of the previous $m$ operators, and $\xi_L^2$ represents the variance of the evaluations of $f_L$. Therefore, for a sufficiently large set of historical data, the we assume that $\xi_L^2$ will be small.
\end{remark}
\smallskip

\begin{remark}
If the low-fidelity is evaluated at all points in $\bm{X}_H$, we see that 
\begin{align*}
    \cov{H}{L}{L} [\cov{L}{L}{L}]^{-1}\cov{L}{H}{L} = \cov{H}{H}{L},
\end{align*}
resulting in a simplification of the upper bound to
\begin{align*}
    \cov{H}{H}{\delta} + \xi_H^2I + \xi_L^2 \cov{H}{L}{L}[\cov{L}{L}{L}]^{-2} \cov{L}{H}{L}.
\end{align*}

Additionally, we see that as the high- and low-fidelity noise terms approach 0, the conditional covariance approaches $\cov{H}{H}{\delta}$. 
This result is a generalization of the simplification found in the proof of Theorem 3.2 in \cite{lau2023multi}, where $\bm{X}_H\subseteq \bm{X}_L$ and $\xi_L^2=0$.
\end{remark}
\smallskip


An upper bound on the conditional covariance allows us to establish an upper bound on the maximum information gain $\gamma_T$, a metric quantifying the greatest amount of information that can be learned after $T$ points of a GP $f$ are sampled. Suppose $f$ is sampled at points $A\subseteq\mathcal{X}$, resulting in a vector of noisy evaluations $\bm{y}_A$ and a vector of true values $\bm{f}_A$. 
Then, denoting the entropy of a vector by $H(\cdot)$, the information gain is defined as $I(\bm{y}_A;\bm{f}_A):= H(\bm{y}_A)-H(\bm{y}_A|f)$, and the maximum information gain is  
\begin{align} \label{eq:info_gain}
    \gamma_T := \underset{A\subset \mathcal{X}, |A|=T}{\max}I(\bm{y}_A; \bm{f}_A).
\end{align}

\smallskip
\begin{lemma}[Info. Gain Bound for a Noisy \ar\ GP] \label{thm:information_gain}
Let $\xi_H^2$ and $\xi_L^2$ be the variance of the high- and low-fidelity measurement noise of a linear auto-regressive GP.
Then the maximum information gain $\gamma_T$ has the upper bound \cite{srinivas2012information}
\begin{align} \label{eq:gamma_tilde}
        \tilde{\gamma}_T := \frac{1/2}{1-e^{-1}}
        \max_{m_1,...,m_T} \sum_{t=1}^{h(T)} \log \left(1{+}\xi_H^{-2} m_t \lambda_t(\tilde{\bm{k}}) \right),
    \end{align}
    where $\sum_{i=1}^T m_i=T$, $h(T) = \min \{T, |\bm{X}_H|\}$, and $\lambda_t(\tilde{\bm{k}})$ are the eigenvalues of the matrix $\tilde{\bm{k}}$ from Lemma \ref{thm:conditional_covariance}. \hfill \qed
\end{lemma}

\smallskip
\begin{remark}
    We see that the bound on the information gain depends on the magnitude of the eigenvalues of $\tilde{\bm{k}}$. As such, we can evaluate the benefit of a multi-fidelity model by comparing the eigenvalues of $\tilde{\bm{k}}$ with the eigenvalues of the single-fidelity covariance $\cov{H}{H}{H}$. When the eigenvalues of $\tilde{\bm{k}}$ are smaller than the eigenvalues of $\cov{H}{H}{H}$, the information gain bound is lower for the \ar\ GP than a single-fidelity GP with the same data.
\end{remark}

\smallskip
Using this bound on the information gain, we now present our main result: a bound on the regret of an \ar\ model.
\begin{theorem}[Regret Bounds for UCB on an \ar]\label{thm:regret}
Let $f$ be a sample function from a linear auto-regressive GP \eqref{eq:ar1} over the discrete domain $\mathcal{X}$. Set $\delta \in (0,1)$ and $\beta_f = 2\log(|\mathcal{X}|t^2\pi^2/6\delta)$.
Then, the points $\{\x_1, \x_2, \dots \x_T\}$ obtained from Algorithm \ref{alg:mf} satisfy with probability at least $1-\delta$,
\[
R_T \leq \sqrt{C_1 T \beta_T \tilde{\gamma}_T}.
\]
Here, $\tilde{\gamma}_T$ is the information gain bound established in Lemma \ref{thm:information_gain} and $C_1 = 8 v_{MF}^2 /\log(1+v_{MF}^2\xi^{-2})$, where $v_{MF}^2$ is the variance of the \ar\ GP, given by $v_{MF}^2 = \rho v_L^2 + v_\delta^2$. \hfill\qed
\end{theorem}

The proof of this theorem closely follows the proof of Theorem 1 in \cite{srinivas2012information}.

\smallskip
\begin{remark}
    The regret of \ucb\ is upper bounded by the information gain. As such, lowering the information gain bound will improve the cumulative regret bound. In particular, when the eigenvalues of $\tilde{\bm{k}}$ are smaller than the eigenvalues of $\cov{H}{H}{H}$, the \ar\ model improves the regret. 
    
    Further, when $f_L$ closely matches $f$, the variance of $\delta(\x)$ decreases together with the eigenvalues of $\tilde{\bm{k}}$. This, in turn, results in a lower regret bound. In other words, when variations between operators have little effect on the HRI performance curve, Alg. \ref{alg:mf} will obtain a very small regret.
\end{remark}

\section{Numerical Simulations}
We conduct two numerical simulations to demonstrate the performance of Algorithms \ref{alg:mf}, \ref{alg:csf}, and \ref{alg:lsf}. First, we apply these algorithms to the undisturbed LTI model \eqref{eq:Z_i_dot}. Then, we show the robustness of our approach by applying it to an LTI system with an unknown disturbance.

\subsection{LTI Model}
Consider the LTI system \eqref{eq:Z_i_dot} with $n=2$ degrees of freedom. Because we model the robot using an impedance model, the end-effector's motion is assumed to be independent in each direction. By letting $M_m = I_2$,  we assume $B_m$ and $K_m$ will also be scalar matrices. Thus, we assume $K$ possesses the structure
\[
K(\x) = 
\begin{bmatrix}
    x_1 & 0 & x_2 & 0 & x_3 & 0 \\
    0 & x_1 & 0 & x_2 & 0 & x_3
\end{bmatrix}.
\]
Henceforth, we use $\x = (x_1, x_2, x_3) \in \mathcal{X}$ as the optimization parameter, where $\mathcal{X}$ is a $11\times 11\times 11$ hyperrectangle with span $x_1 \in [0.25,0.45]$, $x_2 \in [0.85,0.95]$, and $x_3 \in [0.02,0.22]$.

Next, we generate data for $m=9$ previous operators.
For the performance functions, we aim to minimize the human effort by setting $Q=\text{diag}(0.1,0.1,0.1,0.1,10,10)$ and $R=I_2$. We randomly draw $k_d^i\sim N(10, 5)$, $k_p^i\sim N(20, 5)$ and set $K_d^i = k_d^i I_n$, $K_p^i = k_p^i I_n$. 
An initial condition $Z_i(0) = [I_n \quad \bm{0}]^T$ is chosen to model an initial error in position.
The performance $f_i$ from \eqref{eq:f_i} is approximated using a finite integral from $\tau=0$ to $\tau=10$. Each $f_i$ is evaluated for $20$ random sets of $\x \in \mathcal{X}$ with additive Gaussian noise $\eta \sim N(0,10^{-4})$.

\begin{figure}[tb]
    \centering
    \includegraphics[width=0.9\linewidth]{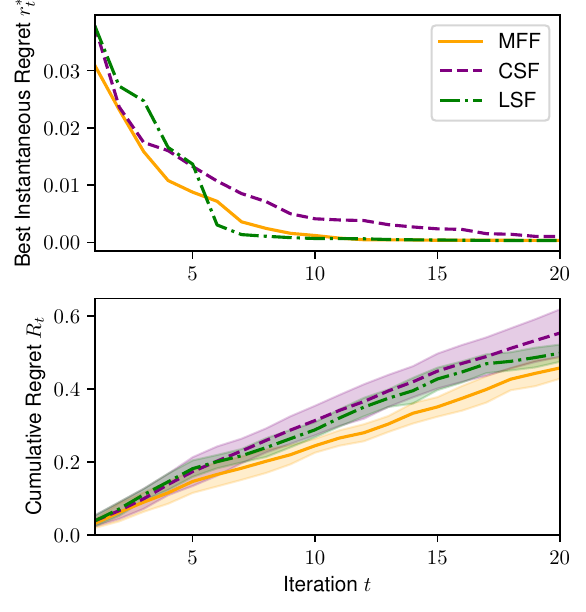}
    \caption{Best Instantaneous and Cumulative Regret (averaged across 20 trials) when \ucb\ is used to select control gains for system \eqref{eq:Z_i_dot} with no disturbance. Error bars represent one standard deviation across 20 Monte Carlo trials.}
    \label{fig:undisturbed}
\end{figure}

We run $20$ Monte Carlo simulations involving the random selection of previous data points and operator gains $K_d$, $K_p$. Fig. \ref{fig:undisturbed} displays the averages of best and cumulative regrets across the simulations. 
We see that MFF leads to a general improvement in the cumulative regret, especially for higher iteration counts. Between the single-fidelity approaches, LSF has a lower regret and tighter variance than CSF.

The best instantaneous regret plot shows that MFF typically makes better selections than CSF or LSF in the first few iterations. After around 10 iterations, LSF and MFF have found a selection with very low regret while CSF fails to find an optimal selection even after the 20 iterations.

These results indicate that data from the previous operators is beneficial when it is incorporated through a multi-fidelity structure. Incorporating previous data through CSF increases the regret compared to ignoring it in LSF.

\subsection{LTI Model with Disturbance}
Because our techniques rely only on input-output data, the technique is inherently robust to deviations in the model.
To demonstrate this, suppose the feedback linearization of \eqref{eq:u_q} is imperfect, resulting in a disturbance affecting the evolution of $\dot{\e}$. Then the disurbed evolution of the system is
\begin{align} \label{eq:disturbed_ss}
    \dot{Z}_i = \mathcal{A}^{i}Z_i + \mathcal{B}^{i} \bm{u} + \bm{d},
\end{align}
where $\bm{d}\in \real^n$ is an unknown but constant disturbance to the system.
Specifically, we model a disturbance on states directly affected by the control input \eqref{eq:u_bar} by setting $\bm{d} = [0, 0, 0.05, 0.05, 0, 0]^T$.

\begin{figure}[tb]
    \centering
    \includegraphics[width=0.9\linewidth]{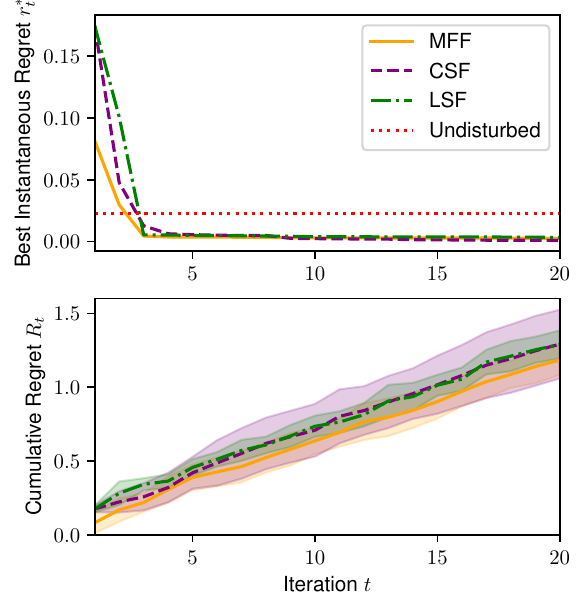}
    \caption{Best Instantaneous and Cumulative Regret (averaged across 20 trials)  when \ucb\ is used to select control gains for the disturbed system \eqref{eq:disturbed_ss}. The dashed red line indicates the regret when the optimal controller from the undisturbed system is used on the disturbed system. Error bars represent one standard deviation across 20 Monte Carlo trials.}
    \label{fig:disturbed}
\end{figure}

We plot the regret from the MFF, CSF, and LSF approaches in Fig. \ref{fig:disturbed}. We also show the regret incurred when the optimal controller from the undisturbed system is used on the disturbed system.

In this case, the disturbance increases the means and spreads of the cumulative regret. Still, on average, MFF performs better than LSF or CSF. 
Additionally, on average, all three algorithms identify a better controller than the optimal undisturbed controller in three iterations.

\section{Conclusion}
We provide a multi-fidelity framework to find the optimal set of impedance parameters for a human-robot cooperative manipulation system using only input-output data. By treating prior operator data as a low-fidelity model, we are able to further optimize the system's performance for a new operator. We establish how the \ar\ model improves the regret bound through the conditional covariance and then numerically simulate human-robot cooperative manipulation to demonstrate this improvement in regret.

In future work, we plan to validate this framework by conducting physical experiments with human subjects and a robotic manipulator.

\bibliographystyle{IEEEtran}
\bibliography{hri_paper.bib}

\end{document}